\documentclass[acmtog, nonacm]{acmart}
\acmSubmissionID{504}

\usepackage{booktabs} 

\usepackage[ruled]{algorithm2e} 

\SetAlFnt{\small}
\SetAlCapFnt{\small}
\SetAlCapNameFnt{\small}
\SetAlCapHSkip{0pt}

\acmJournal{TOG}

\begin{document}
\title{TaleCrafter: Interactive Story Visualization with Multiple Characters}
\author{Yuan Gong}
\affiliation{%
 \institution{Tsinghua Shenzhen International Graduate School, Tsinghua University}\country{China}}
\email{gong-y21@mails.tsinghua.edu.cn}

\author{Youxin Pang}
\affiliation{
    \institution{NLPR, Institute of Automation, Chinese Academy of Sciences}\country{China}}
\email{pangyouxin2020@ia.ac.cn}

\author{Xiaodong Cun}
\affiliation{
\institution{Tencent AI Lab}
\city{ShenZhen}
\country{China}}
\email{vinthony@gmail.com}

\author{Menghan Xia}
\affiliation{
\institution{Tencent AI Lab}
\city{ShenZhen}
\country{China}}
\email{menghanxyz@gmail.com}

\author{Yingqing He}
\affiliation{
\institution{Hong Kong University of Science and Technology}
\city{Hong Kong}
\country{China}}
\email{yhebm@connect.ust.hk}

\author{Haoxin Chen}
\affiliation{
\institution{Tencent AI Lab}
\city{ShenZhen}
\country{China}}
\email{jszxchx@gmail.com}

\author{Longyue Wang}
\affiliation{
\institution{Tencent AI Lab}
\city{ShenZhen}
\country{China}}
\email{vincentwang0229@gmail.com}

\author{Yong Zhang}
\authornote{Corresponding authors. \\ 
Project~: ~\url{https://github.com/VideoCrafter/TaleCrafter}}
\author{Xintao Wang}
\author{Ying Shan}
\affiliation{
\institution{Tencent AI Lab}
\city{ShenZhen}
\country{China}}

\author{Yujiu Yang}
\authornotemark[1]
\affiliation{%
 \institution{Tsinghua Shenzhen International Graduate School, Tsinghua University}
 \country{China}
 }
\email{yang.yujiu@sz.tsinghua.edu.cn}

\makeatletter
\let\@authorsaddresses\@empty
\makeatother


\begin{abstract}
Accurate Story visualization requires several necessary elements, such as identity consistency across frames, the alignment between plain text and visual content, and a reasonable layout of objects in images. Most previous works endeavor to meet these requirements by fitting a text-to-image (T2I) model on a set of videos in the same style and with the same characters, \textit{e.g.,} the FlintstonesSV dataset. However, the learned T2I models typically struggle to adapt to new characters, scenes, and styles, and often lack the flexibility to revise the layout of the synthesized images.
This paper proposes a system for generic interactive story visualization, capable of handling multiple novel characters and supporting the editing of layout and local structure. It is developed by leveraging the prior knowledge of large language and T2I models, trained on massive corpora. The system comprises four interconnected components: story-to-prompt generation (S2P), text-to-layout generation (T2L), controllable text-to-image generation (C-T2I), and image-to-video animation (I2V). First, the S2P module converts concise story information into detailed prompts required for subsequent stages. Next, T2L generates diverse and reasonable layouts based on the prompts, offering users the ability to adjust and refine the layout to their preference. The core component, C-T2I, enables the creation of images guided by layouts, sketches, and actor-specific identifiers to maintain consistency and detail across visualizations. Finally, I2V enriches the visualization process by animating the generated images.
Extensive experiments and a user study are conducted to validate the effectiveness and flexibility of interactive editing of the proposed system.

\end{abstract}
\begin{teaserfigure}
\centering
\includegraphics[width=\linewidth]{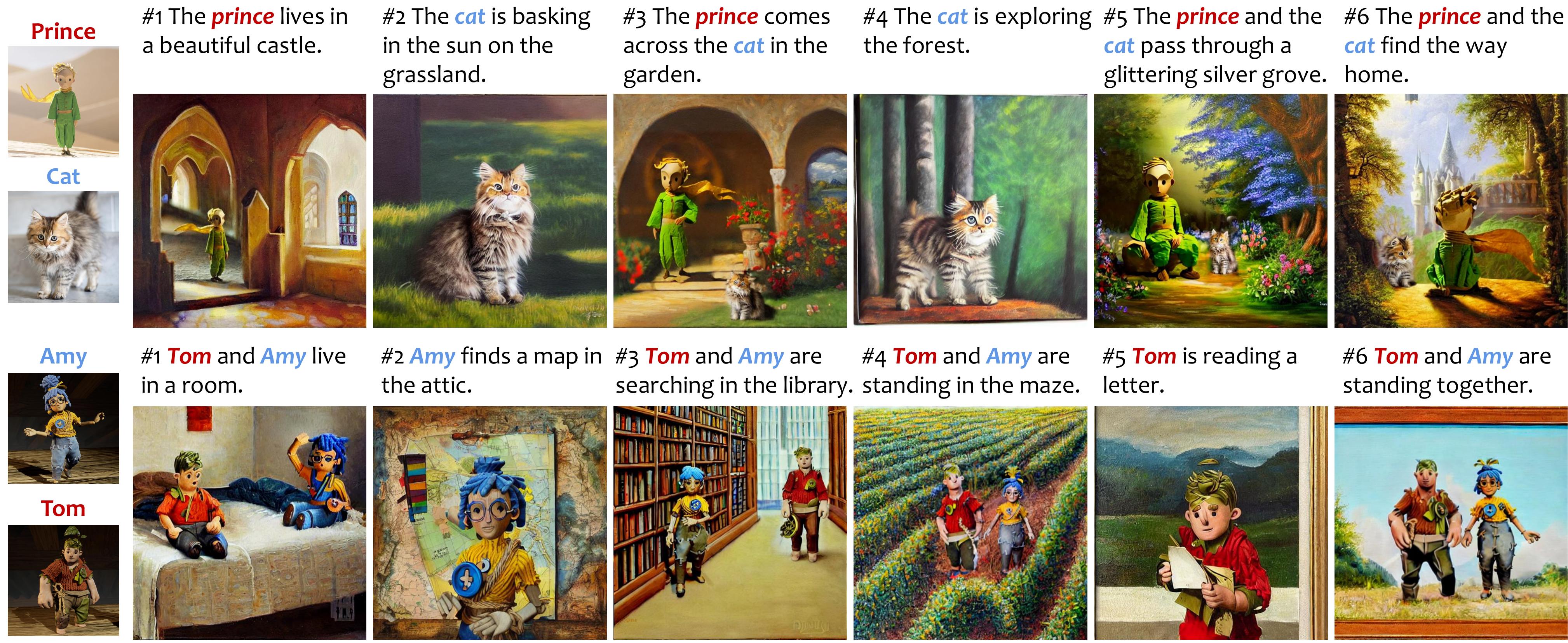}
  \caption{
The visual examples of our story visualization system. Given the story and multiple characters, the S2P component first generates a series of prompts from the story using GPT-4. Then the T2L component creates a reasonable layout given a generated prompt. The core C-T2I component takes multi-modality inputs, such as prompt, layout, and sketch, to render an image with the specified characters, locations, and local structures. Finally, the I2V component animates those generated images. The style is specified by \textit{"oil painting"}. \textbf{Video results can be found in the supplementary materials.}
  }
  \setlength{\abovecaptionskip}{-0.1cm}   
  \label{fig:teaser}
\end{teaserfigure}

%




\maketitle

\newcommand{\todo}[1]{\textcolor{red}{TODO #1}}
\newcommand{\tocite}{\textcolor{red}{TO Cite}}
\newcommand{\modelname}{LVDM\xspace}

\section{Introduction}
\label{sec:intro}

      


Story visualization, also known as visual storytelling, is a vital method for effectively conveying narrative content to a diverse range of audiences. It has a wide range of applications in education and entertainment~\cite{yin2022styleheat}, \textit{e.g.,} children's comic books. In this work, story visualization is formulated as such a problem, \textit{i.e.,} given a story in plain text and the portrait images of a few characters, generate a series of images to express the story visually.

An eligible story visualization should meet several essential requirements to provide an accurate visual representation of a narrative. First, identity consistency. Maintaining consistent depictions of characters and environments across all frames or scenes is crucial. Second, text-visual alignment. The visual content should align closely with the textual narrative, accurately representing the events and interactions described in the story. Third, clear and logical layout. Objects and characters within the generated images should be arranged in a reasonable and logical layout. This organization helps to guide the viewer's attention seamlessly through the narrative, making it easier to understand.

Pioneer works in story visualization typically train models on specific datasets containing characters and styles that are consistent throughout. 
Two popular datasets include PorotoSV~\cite{li2019storygan} and FlintstonesSV~\cite{maharana2021integrating}, which feature cartoon styles and limited character variation. 
Most earlier approaches~\cite{li2019storygan, maharana2021improving, maharana2021integrating, li2022word, chen2022character, maharana2022storydall} rely on GAN or VAE-based methods, incorporating text encoders to project text into a latent space, decoders to generate images conditioned on the text, and image and story-level discriminators to retain visual quality and consistency.
Some studies now leverage diffusion models to capture conditional distribution, often using a pre-trained T2I model for initialization. 
For instance, AR-LDM~\cite{pan2022synthesizing} introduces a latent diffusion model, autoregressively conditioned on historical captions and synthesized images to predict current frames. 
Make-A-Story~\cite{rahman2022make} proposes an autoregressive model with a visual memory module, capturing actor and background context across generated frames for content consistency.

However, these methods encounter two unavoidable limitations.
First, they face challenges in generalizing to new actors and scenes, as they are trained on specific datasets to fulfill the primary two requirements. 
Recent work~\cite{jeong2023zero} investigates the potential of zero-shot story visualization using a pre-trained T2I model to enable adaptation to any new character and scene. 
The process involves generating an image and subsequently replacing the human face with a supplied one. 
Unfortunately, this approach neither accommodates multiple characters nor supports objects apart from the human face.
Second, none of these methods take into account the third requirement, \textit{i.e.}, the layout of the generated image or local object structure, with all information implicitly controlled by the text. 
Although several text-to-image and layout-to-image methods~\cite{zhang2017stackgan,rombach2022high, hong2018inferring,liang2023layout, li2023gligen} incorporate layout as input or intermediate result, their focus lies solely on single image generation rather than story visualization, without considering cross-frame consistency.

In this work, we introduce a versatile interactive story visualization system that satisfies all three requirements, building on the knowledge of large-scale language and text-to-image (T2I) models trained on extensive corpora. 
This system can adapt to various new characters and support layout and local structure editing beyond the capabilities of previous methods. 
Our system consists of four components: story-to-prompt generation (S2P), text-to-layout generation (T2L), controllable text-to-image generation (C-T2I), and image-to-video animation (I2V).
Given a story, S2P leverages a large language model to generate prompts that depict the visual content of images based on instructions, including events, scenes, and characters. 
Subsequently, T2L utilizes the prompt to create an image layout that offers location guidance for the main subjects, while allowing interactive refinement of the layout.
The core component, C-T2I, renders images conditioned on the layout, local sketch, and prompt, while preserving the identity of multiple characters. 
The prompt conveys the image content, whereas the layout and local sketch represent the subjects' locations and detailed local structures, respectively. 
To preserve identity, the model learns a small set of personalized weights for each character. 
C-T2I facilitates interactive editing of local structures and seamless replacement of characters with new ones.
Finally, I2V enriches the visualization process by animating generated images for more vivid presentation.
Visual results are shown in Fig.~\ref{fig:teaser}. 

Our main contributions are in two key aspects:
\begin{itemize}
    \item We propose a versatile and generic story visualization system that leverages large language and pre-trained T2I models for generating a video from a story in plain text. This system can handle multiple novel characters and scenes. 

    \item We develop a controllable, multi-modality text-to-image generation module, C-T2I, which serves as the core component of the visualization system. 
    This module focuses on identity preservation for multiple characters and emphasizes structure control in terms of layout and local structure. 
\end{itemize}


\begin{figure*}
    \centering
    \includegraphics[width=\linewidth]{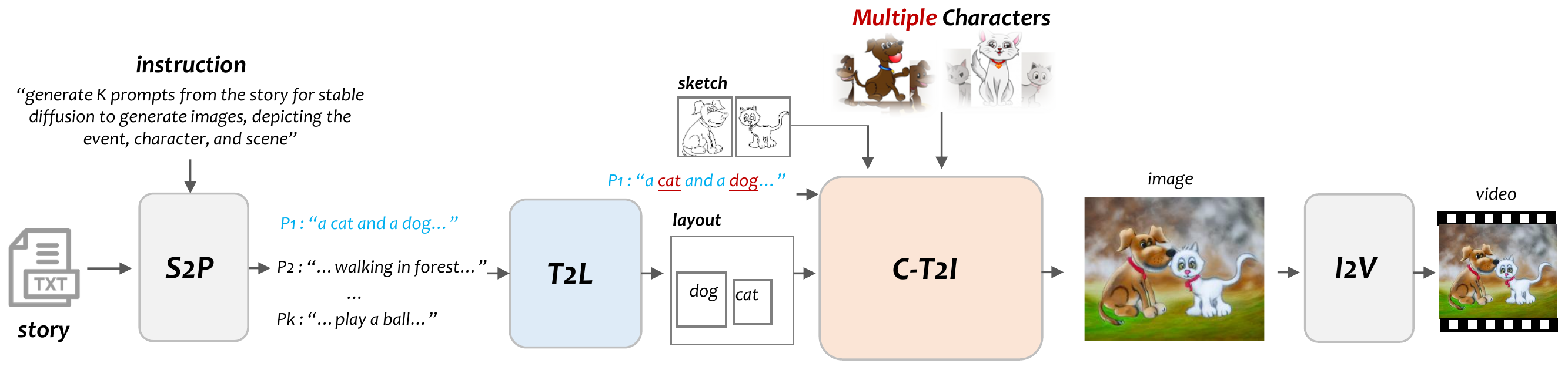}
    \caption{The pipeline of our interactive story visualization system. The system comprises four components. (a) Story-to-prompt (S2P): a large language model is utilized to bridge the gap between the literary and artistic descriptions and the descriptions fed into T2I models. 
It comprehends the content in the given story and converts it into prompts suitable for T2I models, following the given instructions. 
(b) Text-to-layout (T2L): generates a reasonable layout for the main subjects in the prompt. 
(c) Controllable text-to-image (C-T2I): given various conditions such as prompt, layout, sketch, and a few images of each character, generates consistent-character images. It enables interactive editing of character, layout, and local structure through sketches. 
(d) Image-to-video (I2V): extracts depth from the image and converts it into a video by setting the camera path for novel view synthesis. 
   }
    \label{fig:pipeline}
\end{figure*}

\section{Related Work}

\subsection{Story Visualization}

The earlier works in the field of story visualization primarily relied on GAN or VAE-based approaches~\cite{li2019storygan,maharana2021improving,maharana2021integrating,li2022word, chen2022character, song2020character}. 
For instance, StoryGAN~\cite{li2019storygan} uses both the full story and individual sentences as inputs to generate contextually relevant images, employing image and story discriminators. 
DUCO-StoryGAN~\cite{maharana2021improving}, on the other hand, introduces a dual learning framework that utilizes video captioning to enhance semantic alignment between stories and generated images.
Several studies take advantage of the long-range dependence properties of transformers, such as VP-CSV~\cite{chen2022character} and StoryDALL-E~\cite{maharana2022storydall}. 
The latter adapts a pre-trained model to a specific dataset to leverage the prior knowledge.

Recently, diffusion models have shown success in various applications, including image, video, and audio generation~\cite{rombach2022high,ho2022imagen,liu2023audioldm}. Several works integrate diffusion models into story visualization, replacing GANs~\cite{pan2022synthesizing,rahman2022make}. AR-LDM~\cite{pan2022synthesizing} adopts StoryDALL-E's setup for story continuation, and utilizes latent diffusion models as image generators while aggregating information from current and previous prompts using an autoregressive model.
Make-A-Story~\cite{rahman2022make} proposes an autoregressive diffusion-based framework featuring a visual memory module. Similar to AR-LDM, it leverages historical results, albeit relying on the cross attention mechanism and interacting in the feature space, rather than text embedding.

However, these methods often struggle to generalize to novel characters and scenes, as they inherently fit the model to specific datasets like FlintstonesSV. Consequently, the model can only recall the characters and scenes from the training dataset. Aiming to eliminate this limitation, one recent study~\cite{jeong2023zero} focuses on zero-shot story visualization supporting novel characters and scenes, proposing a method for character identity replacement in images using diffusion models—an approach reminiscent of face swapping. However, this method's scope is limited to single human faces, and the identity preservation and consistency across images remain unsatisfactory.
Our method targets zero-shot story visualization as well, supporting multiple novel characters and scenes. 
To ensure identity consistency, we optimize a small set of model weights for each characte and propose a personalized inpainting method to compose multiple characters. 
Moreover, our approach allows control over layout and local object structures, surpassing the capabilities of previous works.

\subsection{Text-to-image Generation}

A significant number of Text-to-Image (T2I) methods are founded upon GANs~\cite{reed2016generative, xu2018attngan, zhang2017stackgan}. Typically, these methods involve a text encoder and an image generator. Some approaches~\cite{hong2018inferring,li2019object,qiao2021r} employ intermediate layout generation to simplify image generation directly from text.
Recently, diffusion models have demonstrated potential in image and video generation, with several studies~\cite{ramesh2022hierarchical, ramesh2021zero, rombach2022high, saharia2022photorealistic} improving image quality and diversity using diffusion models. 
However, these methods mainly concentrate on the alignment between text and a single generated image, without considering identity consistency across multiple images.

Despite the success of inversion methods~\cite{ruiz2022dreambooth,gal2022image,yang2023controllable,shi2023instantbooth} in maintaining identity in diffusion-based T2I generation, they typically excel with single concepts while struggling to cope with multiple concepts. 
Custom Diffusion~\cite{kumari2022multi} aims to compose multiple concepts but falters when dealing with similar-looking concepts, such as cats and dogs.
Since composing multiple concepts or characters in an image using inversion remains a challenge, we propose a controllable T2I model for personalized inpainting, which tackles composition from a different perspective.


\section{Method}
We propose an interactive story visualization system that supports interactive editing of character, layout, and local structure.
Different from most previous works, it can handle consistent generation of multiple characters and generalize to novel characters and scenes. 
As shown in Fig.~\ref{fig:pipeline}, the system comprises four components, \textit{i.e.,} story-to-prompt (S2P), text-to-layout (T2L), controllable text-to-image (C-T2I), and image-to-video (I2V).

\subsection{Story-to-prompt Generation}
The given story could be a brief sentence, \textit{e.g.,} \textit{``a cat and a dog have a wonderful day."}, or it could be a long detailed one with literary techniques. 
Both might not fit the taste of current T2I models that are trained with captions depicting the events, scenes, and objects in images.  
Recently, amazing breakthroughs have been achieved in the development of large language models, such as GPT-4~\cite{openai2023GPT4} and PaLM 2~\cite{anil2023palm}. 
GPT-4 is trained on a massive multi-modality corpus, including vision and language, which is an appropriate tool to bridge the gap between literary descriptions and the descriptions for T2I models.  We use GPT-4 in this work. 

The instruction matters in GPT-4. 
The basic elements in the description for T2I models are event, scene, and object. 
Hence, for a given story, we use the instruction, like \textit{``generate K prompts from the story for Stable Diffusion to generate images, depicting event, character, and scene."}
Leveraging the capability of pre-trained T2I models, we exploit key words in text to control the style, \textit{e.g.,} using \textit{``in oil painting style"} as a suffix of the prompts. 

Let $\mathcal{S}$ denote a story in plain text. 
Let $R$ and $F$ denote the instruction and style, respectively. 
The function of the S2P component can be defined as 
\begin{equation}
    [p_1, p_2, ..., p_K] = S2P(\mathcal{S}, R, F, K), 
\end{equation}
where $K$ is the number of prompts to generate. $p_i$ is the $i$-th prompt.

\subsection{Text-to-layout Generation}
Transformers are the most widely used techniques to capture layout distribution~\cite{gupta2021layouttransformer,jiang2022coarse}.
However, autoregressive decoders are revealed to be inflexible for handling partial inputs~\cite{kong2022blt} due to its fixed generation order. 
Recently, discrete diffusion models~\cite{austin2021structured, gu2022vector} are introduced by LayoutDM~\cite{inoue2023layoutdm} for layout generation. 
It achieves satisfying performance and allows various constraints. 

Following LayoutDM, we exploit discrete diffusion models for text-to-layout generation. 
Let $\mathcal{L}=\{B_i\}_{i=1}^{N}$ denote the layout with $N$ objects, where $B_i = (\mathbf{b}_i, l_i)$. 
$\mathbf{b}_i = (x_i,y_i,w_i,h_i) \in \{1,...,M\}^4$ represents the bounding box with $(x_i,y_i)$ as the center and $(w_i,h_i)$ as the width and height. 
The coordinates are normalized and quantized, and $M$ is the number of bins. 
$l_i \in \{1,2,...,C\}$ is the object category. $C$ is the number of object categories. 
The flattened discrete vector $\mathcal{L} = \{x_1,y_1,w_1,h_1, l_1, x_2, y_2,...\}$ is treated as the latent variable with a variable length. 

Following the definition of discrete diffusion models in D3PM~\cite{austin2021structured},  
the forward diffusion process for a discrete scalar with $D$ categories at timestep $t$, $z_t \in \{1,2,...,D\}$, can be defined as: 
\begin{equation}
    q(z_t|z_{t-1}) = \mathbf{v}(z_t)^T \mathbf{Q}_t \mathbf{v}(z_{t-1})^T, 
\end{equation}
where $\mathbf{v}(z_t)$ is one-hot vector of $z_t$ and $\mathbf{Q}_t \in [0,1]^{D \times D}$ is the transition matrix. 
When $\mathbf{z}_t\in \{1,2,...,D\}^N$ is a vector, the forward process is applied to each of its element independently.  
Similar to DDPM~\cite{ho2020denoising}, the reverse denoising process is estimated by a network, \textit{i.e.,} $p_\theta(\mathbf{z}_{t-1}|\mathbf{z}_{t}) \in [0,1]^{N \times D}$. $\theta$ denotes the parameters of a bidirectional Transformer. 
DP3M decomposes $p_\theta(\mathbf{z}_{t-1}|\mathbf{z}_{t})$ as follows and learns to estimate $\tilde{p}_\theta(\mathbf{z}_{0}|\mathbf{z}_{t})$ instead:
\begin{equation}
    p_\theta(\mathbf{z}_{t-1} | \mathbf{z}_{t})  \propto \sum_{\mathbf{z}_0} 
    q(\mathbf{z}_{t-1} | \mathbf{z}_{t}, \mathbf{z}_{0}) \tilde{p}_\theta(\mathbf{z}_{0}|\mathbf{z}_{t}), 
\end{equation}
where $q(\mathbf{z}_{t-1} | \mathbf{z}_{t}, \mathbf{z}_{0})$ has a closed-form solution according the definition of the diffusion process. 
As the flattened $L$ has a variable length, we exploit the padding trick of LayoutDM to handle it. 

We use the training objective in D3PM, including a widely used variational lower bound ${L}_{\text{vb}} $ and an extra loss, \textit{i.e.,}
\begin{equation}
    {L}_{\text{s2p}} = {L}_{\text{vb}} +  \lambda \mathbb{E}_{q(\mathbf{z}_0)} \mathbb{E}_{q(\mathbf{z}_t | \mathbf{z}_0)} [- \log { \tilde{p}_{\theta}(\mathbf{z}_{0}|\mathbf{z}_{t}) }], 
\end{equation}
where $\lambda$ is a trade-off hyperparameter. 

Since the purpose of T2L is to generate a layout according the given text, the above formulation cannot be directly applied. 
We use a language processing tool~\cite{texsmart2021} to extract nouns from the text and treat them as the target objects.
To convert the nouns into categorical labels, we use the class names in the Object365 dataset~\cite{shao2019objects365}. 
As Object365 provides the bounding boxes and categorical labels of objects in image, we use it to train our T2L model.  
The function of the T2L component can be represented as 
\begin{equation}
    \mathcal{L}=\{B_i\}_{i=1}^{N} = T2L(p), 
\end{equation}
where $p$ is the generated prompt from the S2P component. 

\begin{figure}
    \centering
    \includegraphics[width=\linewidth]{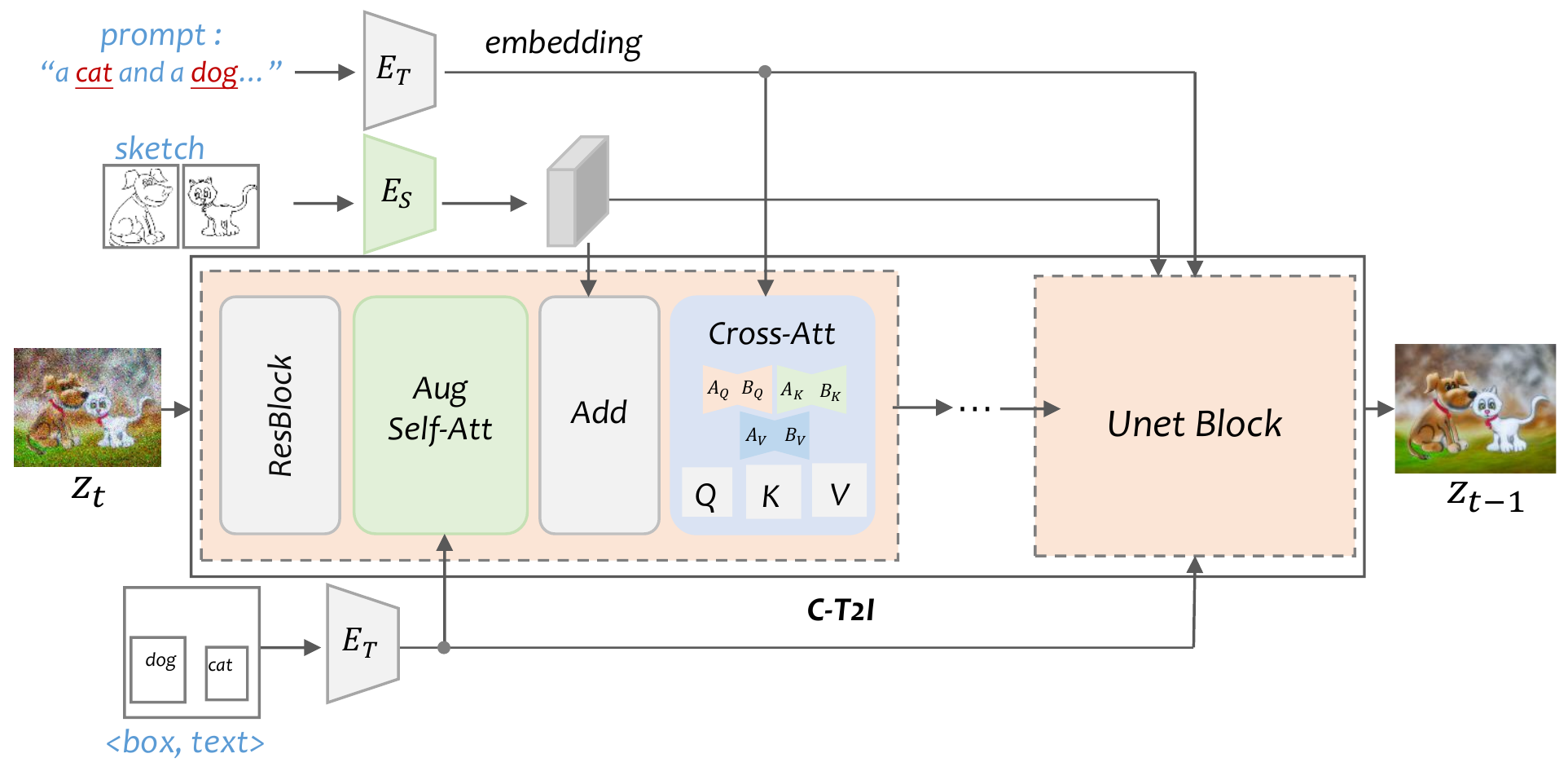}
    \caption{The structure of the C-T2I component. It takes a noisy image as input and generates an image through a single denoising step, conditioning on multiple types of guidance, including prompt, sketch, and bounding box with description. 
For identity consistency, we use LoRA to learn the personalized weights in self and cross-attention layers as well as a specific token for each character. }
    \label{fig:unet}
\end{figure}

\subsection{Controllable Text-to-image Generation}
C-T2I is the core component of the story visualization system (see Fig.~\ref{fig:unet}), which has multi-modality inputs and generates an image with multi-level controls, including the identity, location, and local structure.  
Though many works focus on individual tasks of text-to-image~\cite{rombach2022high}, layout-to-image~\cite{li2023gligen}, sketch-to-image~\cite{voynov2022sketch}, and identity-preservation~\cite{kumari2022multi}, they cannot simultaneously handle the multi-level controls that are essential capabilities of interactive story visualization. 
For example, ~\cite{li2023gligen} is a layout-to-image method with the guidance of text. 
It can locate the objects with input boxes, but it has no control over the local structure and identity. 
\cite{chen2023trainingfree} can specify the location of one specified object, but cannot handle multiple objects.  
~\cite{kumari2022multi} attempts to put two objects in an image, but cannot specify their locations and poses.

Inspired by the latent diffusion model (LDM)~\cite{rombach2022high}, our multi-modality conditional generation model is learned in the latent space, and the distribution is captured using diffusion models.
The structure of C-T2I is shown in Fig.~\ref{fig:unet}. 
To allow the injection of multiple types of input, we modify the structure of the original UNet in LDM. 
In each UNet block, we upgrade the self and cross-attention block and additionally introduce an addition block. 

\paragraph{Identity Preservation.} We exploit the pre-trained CLIP~\cite{radford2021learning} as the text encoder $E_T$ to map the input prompt to an embedding. 
The embedding is fed into the cross-attention module to interact with spatial features. 
For identity preservation, unlike~\cite{ruiz2022dreambooth} and~\cite{kumari2022multi}, we use LoRA to update additional low-rank weights in the self and cross-attention layers instead of fine-tuning all the parameters of a learned T2I model. 
Note that LoRA is only applied to the MLPs of the query, key, and value mappings. 
It can alleviate the overfitting issue and concept forgetting because the original weights are retained, and only a very small set of parameters are learned.  
Given a few images of a character, a token, and a set of LoRA weights are trained specifically. 
The weights are not optimized for multiple characters jointly
because the success rate of composing two characters is not satisfying, especially for characters with similar appearance. 

Let $\mathbf{W} \in \mathbb{R}^{d \times k}$ denote the parameters of a linear mapping, where $d$ and $k$ are the dimensions of the input and output vectors, respectively. 
Let $\mathbf{h}=\mathbf{W} \mathbf{x}$ denote the output vector. $\mathbf{x} \in \mathbb{R}^{k}$ is the input vector. 
Using LoRA to remember a new concept, we learn two low-rank matrices $\mathbf{A} \in \mathbb{R}^{d \times r}$ and $ \mathbf{B} \in \mathbb{R}^{r \times k}$ instead of updating $W$, where $r  \ll \text{min}(d,k)$. 
$\mathbf{A}$ and $\mathbf{B}$ have many fewer parameters than $W$ when $r$ is small. 
The forward pass can be rewritten as: 
$\mathbf{h}=\mathbf{W} \mathbf{x} + \mathbf{B}\mathbf{A}\mathbf{x}$.

\paragraph{Object Localization.}
In the text-to-layout component, the layout contains the coordinates and category.  
In the C-I2I component, we replace the category with a phrase that depicts the object while injecting the learned character token. 
For example, we replace \textit{``dog"} with \textit{``\texttt{sks} dog"} for personalization where \texttt{sks} is the learned token. 
Inspired by \cite{li2023gligen}, the text embedding is extracted using CLIP, while the coordinates are encoded using the Fourier embedding~\cite{mildenhall2021nerf}. 
The two embeddings are concatenated, go through an MLP layer for information alignment, and then fed into the augmented self-attention module that comprises two self-attention layers. 
One is a typical self-attention layer that contains only the interaction among visual features. 
The other contains the interaction between visual features and the location embedding to inject object location.

Let $\mathbf{f}$ denote the visual features. 
Let $\mathbf{e_g} = [E_T(p), \text{Fourier}(\mathbf{b})]$ denote the concatenated embeddings of grounding text $p$ and box $\mathbf{b}=(x,y,w,h)$. 
The first self-attention can be written as:
$\mathbf{f} \leftarrow  \mathbf{f} + \text{SA}(\mathbf{f})$, 
where $\text{SA}(\cdot)$ denotes the self-attention operation. 
The second gated self-attention can be written as : 
$\mathbf{f} \leftarrow \mathbf{f} + \text{tanh}(\alpha) \cdot \text{TS}(\text{SA}([\mathbf{f},\mathbf{e}_g]))$ , where $\alpha$ is a variable with $0$ as the initialization. $\text{TS}(\cdot)$ selects the visual features after the interaction. 

\paragraph{Local Structure Control.}
Almost all current story visualization works do not take the local structure control of objects into consideration. 
The layout and object structure are implicitly determined solely by the text.  
To introduce the flexibility of structure control and interactive editing, we use a sketch as one input. 
Inspired by T2I Adapter~\cite{mou2023t2i}, we use a visual encoder $E_S$ to map the input sketch into the feature space of the UNet. 
The encoder is a stack of four residual blocks~\cite{he2016deep}.  
The predicted sketch features are combined with visual features by an addition module. Note that we first translate and resize the input sketch $I_S$ on a blank canvas according to its corresponding bounding box. 
Then, the created sketch image $\tilde{I}_S$ is fed into $E_S$. 
$\mathbf{f}_s = E_S (\tilde{I}_S)$ denotes the features extracted by $E_S$. The local structure control is realized by the addition of the visual features $\mathbf{f}$ and the generated features $\mathbf{f}$, \textit{i.e.,} $\mathbf{f} \leftarrow \mathbf{f} + \beta \mathbf{f}_s$, where $\beta$ is a parameter to control the strength of applying the structure constraint. It is set to $\beta= 1$ for training, while it can be tuned during inference. 
Note that when $\beta= 0$ for inference, this means the sketch input is not required.

\paragraph{Iterative Generation}
As each character has its own personalized weights, during inference for multiple characters, their tokens and LoRA weights are iteratively applied along with the modules for other modalities. 
For example, given the text \textit{``a dog and a cat in a forest"}, and the boxes and sketches of \texttt{dog} and \texttt{cat}, we first generate an image with the personalized weights of \texttt{dog} and the modified text \textit{``a <sks> dog and a cat in a forest"}. 
\textit{``<sks>"} is the learned token for \texttt{dog}. 
The text of the \texttt{dog} box is set to \textit{``<sks> dog"}. 
Since we have the box of \texttt{cat}, we then inpaint the content of the box with the text \textit{``<yty> cat"}. 
\textit{``<yty>"} is the learned token of \texttt{cat}. 
The inpainting model is a variant of C-T2I with augmented inputs. 
The difference is that we concatenate the noisy image, the original image, and the region mask of the box to form the input with 9 channels. 
The training procedure is the same as C-T2I. 

\paragraph{Training Objective}
Following LDM~\cite{rombach2022high}, we use the variational lower bound for training, 
\begin{equation}
    L_{\text{C-T2I}} =  \mathbb{E}_{z \sim E_I(x), \epsilon \sim \mathcal{N}(0,1), t} \left[ || \epsilon - \epsilon_{\theta} (z_t, t, \mathcal{C})||_2^2   \right], 
\end{equation}
where $x$ here represents an image and $E_I(\cdot)$ is the image encoder that projects image into latent space. 
$\epsilon$ is the sampled noise while $\epsilon_{\theta}(\cdot)$ is the predicted noise. 
$t$ is the timestep and $z_t$ is the noisy image. 
$\mathcal{C} = \{ E_T(p), E_S(\tilde{I}_S), E_T(\mathcal{L})\}$ represents the embeddings of the conditions, \textit{i.e.,} prompt, sketch, and layout. 

\subsection{Image-to-video Generation}
To make story visualization vivid, we introduce an image-to-video component into the system. 
In this component, we mainly focus on the camera movement to generate a video with considering the image depth. 
We exploit a 3D photography method~\cite{shih20203d} to extract the depth and synthesize images under novelty views, which can enhance stereognosis detail and make the scene more realistic than a static image.
This approach allows setting the camera path for various effects, such as zoom-in, circle, and swing. 

\section{Experiments}
\subsection{Settings}

\begin{figure*} [t]
    \centering
    \includegraphics[width=\linewidth]{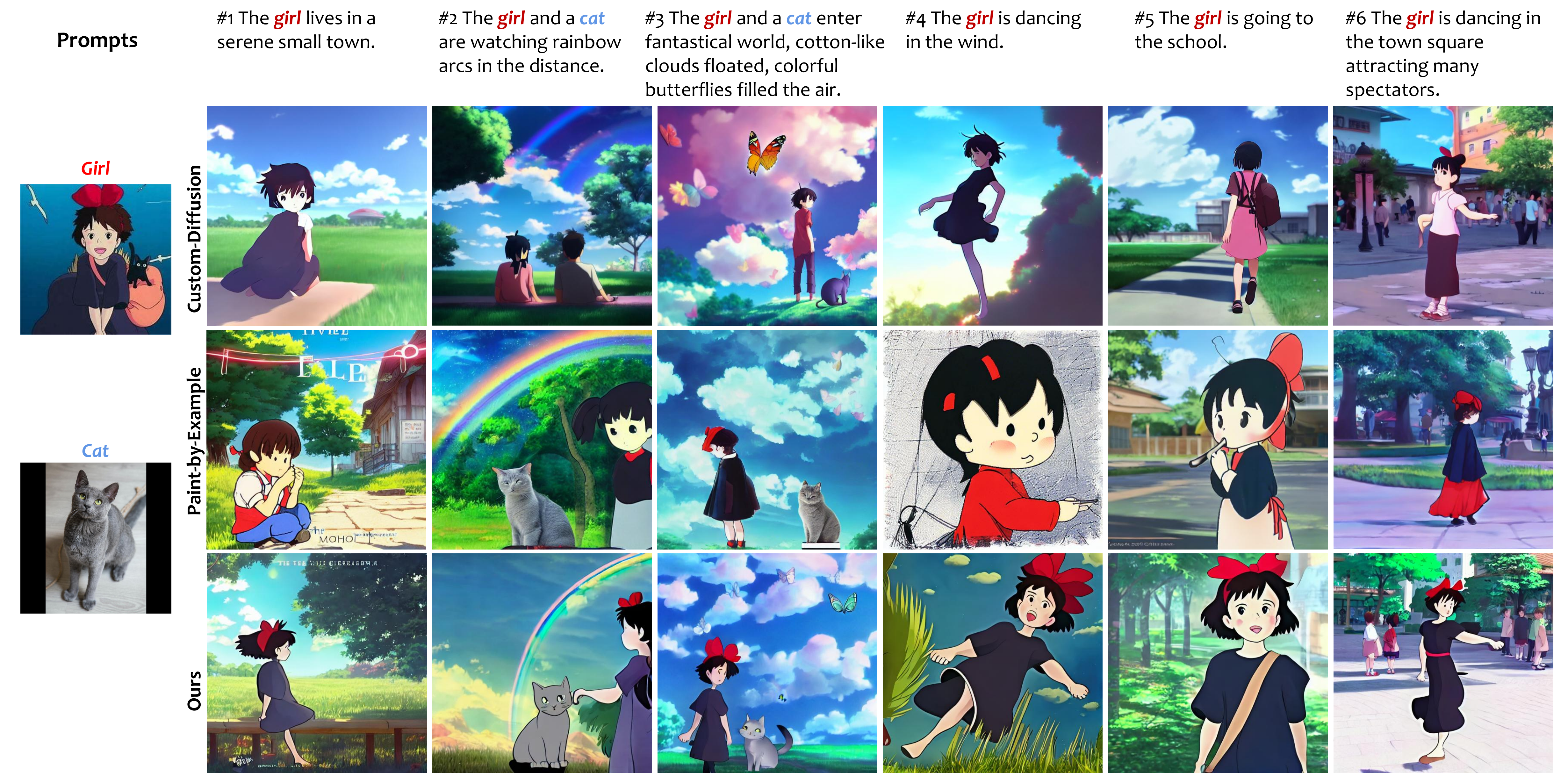}
    \caption{Comparison with Custom-Diffusion and Paint-by-Example. 
    One character is an Anime character and the other is a real cat. 
    The style is specified by \textit{``Ghibli"} for all the three methods. }
    \label{fig:result1}
\end{figure*}

\paragraph{Datasets} 
For the training of the T2L component, we use the Object365 dataset~\cite{shao2019objects365} that contains 365 classes, 2 million images, and 30 million bounding boxes. 
For the C-T2I component, we use the pre-trained Stable Diffusion (v1.4)~\cite{rombach2022high} on LAION-5B~\cite{schuhmann2022laion} as the prior model, including the CLIP encoder, the image auto-encoder, and the diffusion model. 
The sketch encoder and the augmented self-attention module is trained on the Flickr dataset~\cite{plummer2015flickr30k}. The sketches are extracted by PiDiNet~\cite{su2021pixel}. 
To reduce the training difficulty of these two parts, we use the encoder weights of T2I Adapter~\cite{mou2023t2i} and gated self-attention weights of GLIGEN~\cite{li2023gligen} as initialization for training. 
Then, based on the resulting model, we train the personalized LoRA weights for each character with the given 5-9 images. 
For the story-to-prompt generation, we use the large language model GPT-4. 

\paragraph{Evaluation Metrics} 
We evaluate our method along two dimensions. First, we employ text-image similarity in the CLIP feature space to appraise the text alignment of generated images~\cite{hessel2021clipscore}. Subsequently, to gauge the consistency of characters, we utilize image-image similarity in the CLIP image feature space~\cite{gal2022image}. Besides, we also conduct a human preference study for evaluation.

\paragraph{Baselines} 
We compare our method with three approaches: Custom-Diffusion~\cite{kumari2022multi}, Paint-by-Example~\cite{yang2022paint}, and Make-a-Story~\cite{rahman2022make}. 
Custom-Diffusion employs a fine-tuning technique for T2I models and enables joint training for multiple concepts and their composition within a single image. 
Paint-by-Example constitutes an exemplar-guided image editing method, which facilitates character insertion into images for storytelling purposes.
Make-a-Story introduces an autoregressive deep generative framework designed to create stories that exhibit enhanced character and background consistency. However, due to its reliance on extensive story data for training and incompatibility with our limited dataset, we conduct only qualitative experiments.

\paragraph{Implementation Details} 
Both our model and Custom-Diffusion are trained using the same target dataset and regularization dataset, with learning rates of 1e-4 and 1e-5, respectively. 
For each story, we jointly train Custom-Diffusion on two characters. 
Considering that Paint-by-Example constitutes a zero-shot image editing method, we use target images from the training set to edit on identical background images and bounding boxes. 
To ensure fairness, we omit sketch guidance when comparing with them.

\begin{figure}
    \centering
    \includegraphics[width=\linewidth]{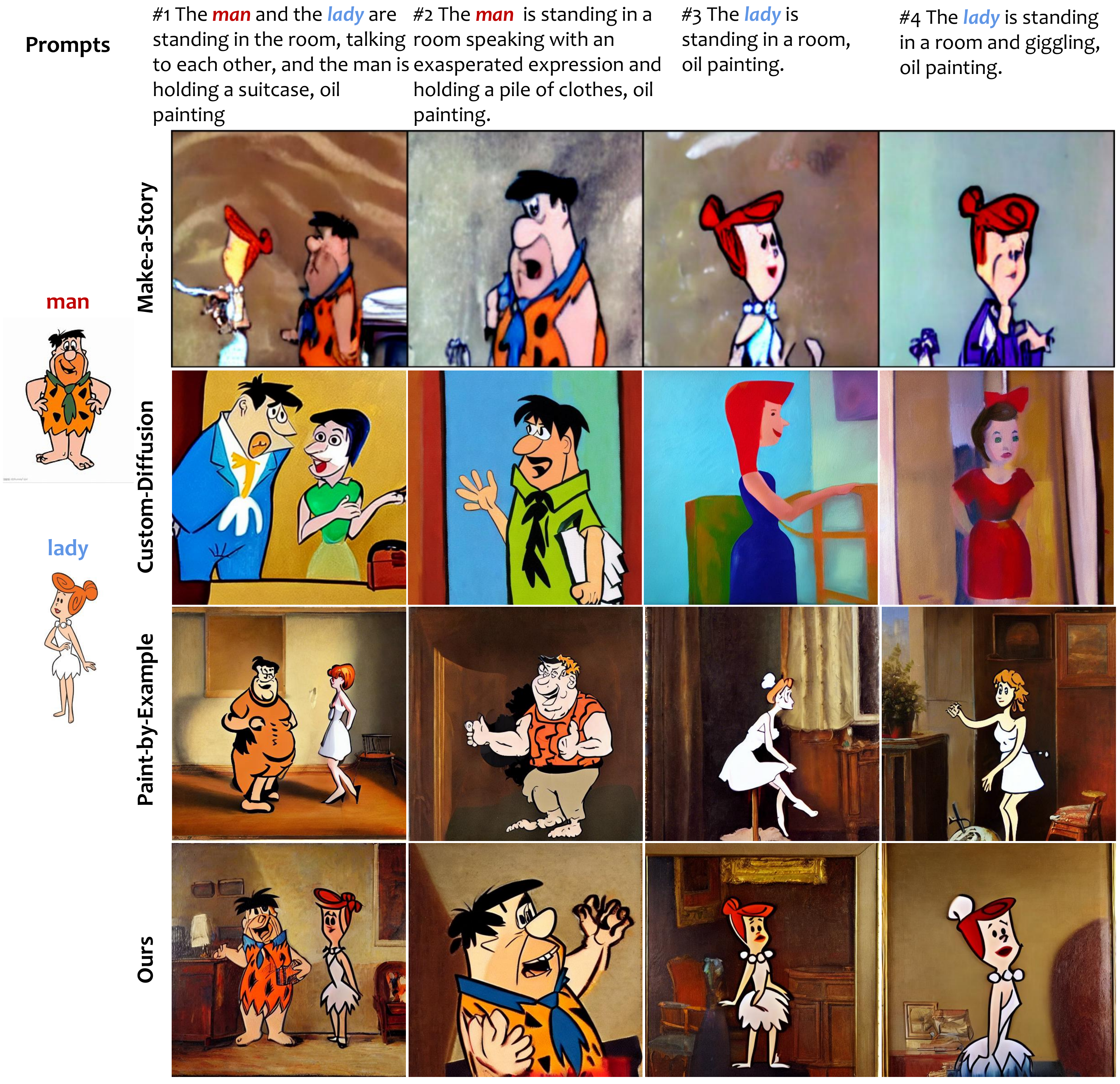}
    \caption{Comparisons with Make-a-Story, Custom-Diffusion, and Paint-by-Example using characters from the FlintsonesSV dataset.}
    \label{fig:result2}
\end{figure}

\subsection{Qualitative Comparisons.} 
In Fig.~\ref{fig:result1} and Fig.~\ref{fig:result2}, we furnish a qualitative assessment of our proposed method compared to the state-of-the-art baselines. 
To compare with Make-a-Story, we utilize examples from the Make-a-Story paper to generate content using the same prompts and characters. 
The results are shown in Fig.~\ref{fig:result2}. 
Our method has higher image quality with fewer artifacts, while better preserving the characters' identities. 
It can be observed that the lady's dress changes across images in the results of Make-a-Story. 
While our results are more consistent. 
Besides, as our method is a generic method that does not require training on a specific data such as FlintstonesSV, it can generate novel scenes out of the range of FlintstonesSV. 

Fig.~\ref{fig:result1} shows the story visualization results of an Anime girl and a real cat.  
Paint-by-Example and Custom-Diffusion perform poorly in identity preservation. 
They always miss the kerchief or generate an inaccurate one. 
Besides, Paint-by-Example fails to convert the cat into an Anime cat according to the image style.  
Custom-Diffusion has a low probability of composing the two characters, as in the second column. 
Differently, our method performs better in preserving identity cross images and can also harmoniously compose characters based on the image style.

\begin{table}[]
\scalebox{0.9}{
\begin{tabular}{llll}
\hline
{Method} & Custom-Diffusion & Paint-by-Example & Ours \\ \hline
text-image sim.   & 0.7422                 & 0.7087                 & \textbf{0.7676}     \\
image-image sim.  & 0.6323                 &  0.6104                &  \textbf{0.6758}    \\ \hline
\end{tabular}
}
\caption{Quantitative comparisons. The text-image and image-image similarity are computed in the CLIP feature space.}
\label{tab:compare}
\vspace{-5mm}
\end{table}

\begin{table}[]
\scalebox{0.9}{
\begin{tabular}{llll}
\hline
{Method} & Custom-Diffusion & Paint-by-Example & Ours \\ \hline
Correspondence   & 1.561                 & 1.566                 & \textbf{2.873}     \\
Coherence        & 1.852                 & 1.498                 & \textbf{2.651}     \\
Quality          & 1.661                 & 1.614                 & \textbf{2.725}     \\ \hline
\end{tabular}
}
\caption{User study on text-image alignment, identity preservation, and image quality. Higher score indicates better performance. }
\label{tab:user-study}
\vspace{-5mm}
\end{table}

\subsection{Quantitative Comparisons.} 
We evaluate on 5 stories, encompassing 35 prompts with 20 samples per prompt, yielding a total of 700 generated images. 
We employ DDIM sampling consisting of 50 steps, and a classifier-free guidance value of 6 across all approaches. 
The text-image similarity in CLIP feature space is used to measure the alignment between the prompt and the generated image. 
The image-image similarity measures the performance of identity preservation. 
We compute the average embedding of the given character images and then compute the similarity between it and generated images. 
As demonstrated in Table~\ref{tab:compare}, our method surpasses the competing methods. 
Securing the highest text-image and image-image similarity, our method demonstrates an enhanced accuracy in generating stories.

\subsection{User Study} 
We perform a user study on the visualization results of 9 stories by Custom-Diffusion, Paint-by-Example, and ours. 
Each story has 3 prompts, resulting in 3 generated images. 
For each given story, 50 participants are asked to rank the performance of the three competing methods from the following three aspects.   
First, whether the generated image accurately reflects the input text description. 
Second, whether the images consistently preserve the character identities. 
Third, the visual quality of the generated images. 
The score ranges from 1 to 3. 
A higher score indicates better performance. 
The average scores in the three aspects are shown in Table~\ref{tab:user-study}. 
In general, our approach achieves the highest scores in all aspects, demonstrating the efficacy of our proposed framework.

\begin{figure}
    \centering
    \includegraphics[width=\linewidth]{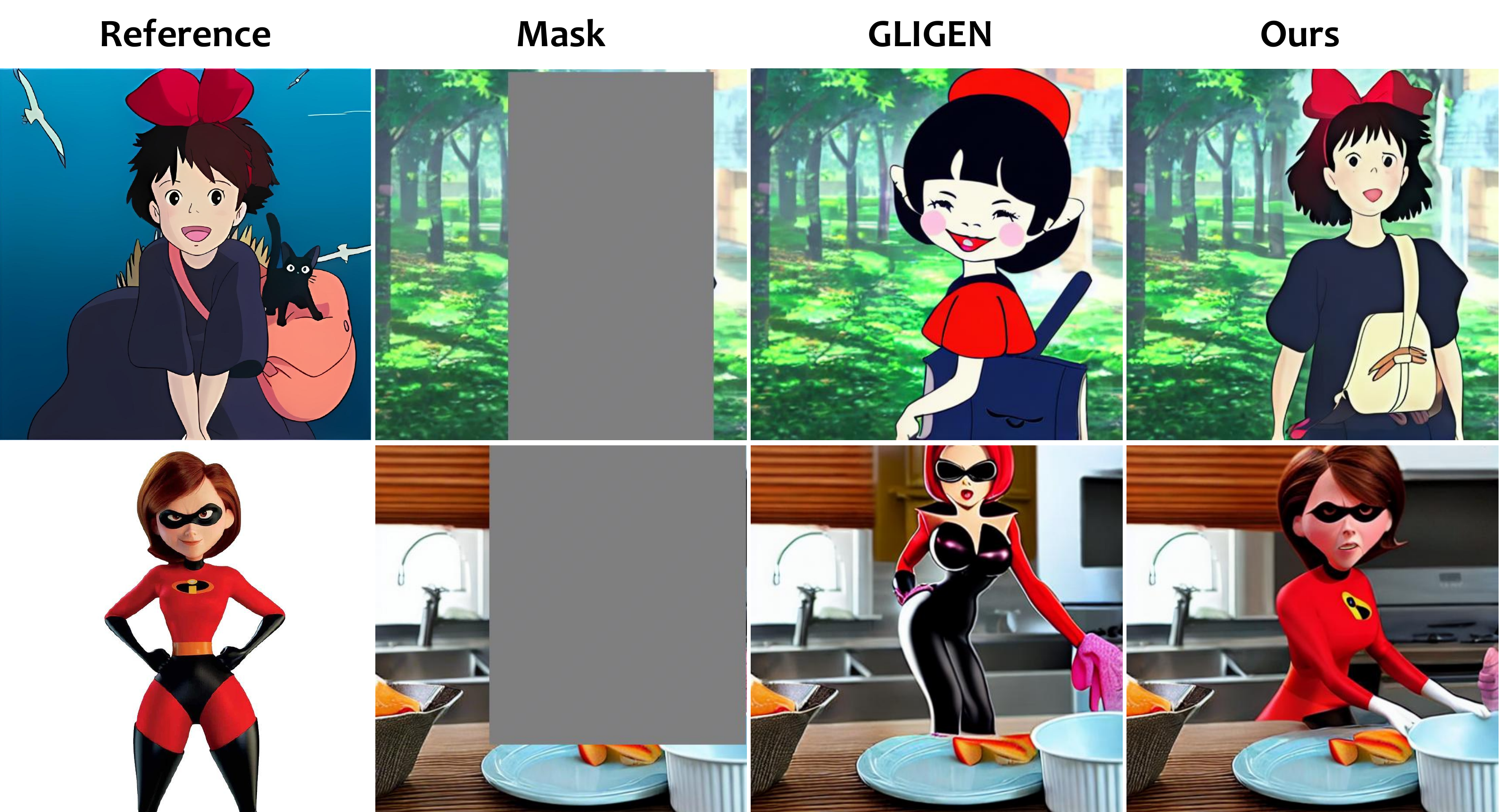}
    \caption{Comparison with GLIGEN on identity control.}
    \label{fig:gligen}
\end{figure}
\begin{figure}
    \centering
    \includegraphics[width=\linewidth]{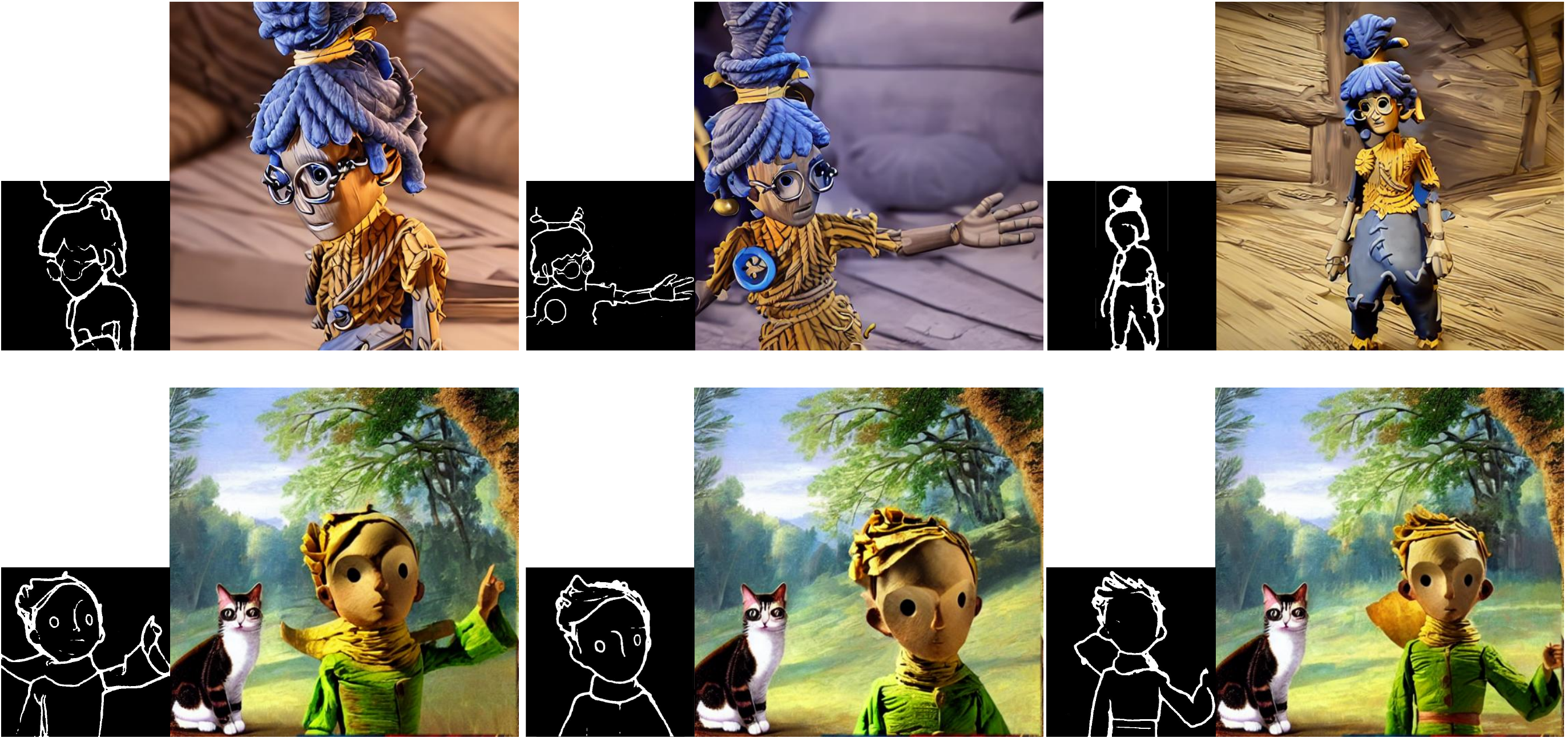}
    \caption{Visualization of sketch controlling. Our model can generate images with the control on local object structure based on the input sketch.}
    \label{fig:sketch}
\end{figure}


\subsection{Interactive Editing and Image Animation} 
\paragraph{Interactive Editing} 
Our system allows the interactive editing of layout, character, and local structure. 
Here, we present the control capability of character and local structure. 
As shown in Fig.~\ref{fig:gligen}, we present a comparison of the identity preservation in GLIGEN. 
GLIGEN takes the reference images as input to fill the masked region while our method uses the personalized token and weights. 
It can be observed that our generated character resembles the reference more than GLIGEN. 
The verification of the structure control is shown in Fig.~\ref{fig:sketch}. 
Given different sketches, the synthesized characters are under the corresponding poses and gestures. 

\paragraph{Image Animation} 
Our I2V component converts an image to a video by extracting depth from the image and setting a camera path. 
We also use text-to-speech to convert the story to audio and combine it with the generated video. 
\textbf{Video results are presented in the supplementary}. 


\section{Limitations}
Our system builds on the pre-trained Stable Diffusion. 
The quality of synthesized images heavily relies on the capability of the pre-trained model. 
Since Stable Diffusion (v1.4) performs poorly in face generation, especially when the face covers only a small region in the image, our system inherits this drawback.      
Another limitation is that the sketch needs to be provided currently. 
The source sketch can come from image retrieval, drawing, 3D rendering, or T2I generated image.
Automatic sketch generation based on reference images and text can be treated as our future work.

\section{Conclusion}
We present an innovative system for generic interactive story visualization capable of handling novel characters and scenes while maintaining identity consistency, alignment between text and visual content, and reasonable object layouts. 
The system's four interconnected components - story-to-prompt generation (S2P), text-to-layout generation (T2L), controllable text-to-image generation (C-T2I), and image-to-video animation (I2V) - work harmoniously to create an interactive and flexible visualization solution. 
Extensive experiments and a user study have demonstrated the effectiveness of the proposed system in story visualization. 

\bibliographystyle{ACM-Reference-Format}
\bibliography{reference}

\appendix

\end{document}